\documentclass[preprint,12pt]{elsarticle}

\usepackage{amssymb}
\usepackage{amsmath}

\usepackage[usenames]{color}
\newcommand{\blue}[1]{\textbf{\textcolor{blue}{#1}}}

\usepackage{multirow}
\usepackage{url}
\usepackage{tabularx}
\usepackage{xltabular}
\usepackage{booktabs}
\usepackage{hyperref}

\journal{Forensic Science International}

\begin{document}

\begin{frontmatter}





\title{CrimeNER: Curated Dataset and Methods for Research and Investigation in Language Intelligence}




\author{Miguel Lopez-Duran} 
\author{Julian Fierrez}
\author{Aythami Morales}
\author{Ruben Tolosana}
\author{Daniel DeAlcala}
\author{Gonzalo Mancera}
\author{Javier Irigoyen}
\author{Oscar Delgado}
\author{Francisco Jurado}
\author{Alvaro Ortigosa}
\affiliation{organization={BiometricsAI},
            addressline={Universidad Autónoma de Madrid (UAM)},
            city={Madrid},
            country={Spain}}
\begin{abstract}

The extraction of critical information from crime-related documents is a crucial task for law enforcement agencies. The extraction of this information can be interpreted as a Named-Entity Recognition (NER) task. However, there is a considerable lack of adequately annotated data on general real-world crime scenarios. To address this issue, we present CrimeNER, a case study of crime-related NER, and a general crime-related Named-Entity Recognition database (CrimeNER-db), consisting of more than 1.5K annotated documents extracted from public reports of terrorist attacks and the US Department of Justice's press notes. We define 4 coarse types of crime entity and 21 fine-grained entity types. We address the quality of the presented database with fully supervised experiments using flat and nested NER architectures and zero- and few-shot experiments using Large Language Models (LLMs). The database is available on GitHub.\footnote{{https://github.com/BiometricsAI/CrimeNER}}
\end{abstract}


\begin{highlights}
\item Database for NER in general crime domain with more than 1.5K annotated documents.
\item Benchmark supervised classical and transformer-based NER  models on CrimeNER-db.
\item Assess LLM performance on crime information extraction with few-shot experiments.
\end{highlights}

\begin{keyword}
Crime Analysis \sep NER \sep Forensics NER \sep Crime NER.


\end{keyword}

\end{frontmatter}


\section{Introduction}
\label{sec::introduction}

Law enforcement agencies are increasingly required to extract and process information from crime-related documents. However, the number of documents that need to be processed is increasing rapidly, so manually extracting these data is not feasible~\cite{madikeri2025autocrime}. In addition, manually annotating sufficient data to train reliable models from scratch is really time consuming and inefficient. 

Automatic extraction of criminal information from document repositories (including web/HTML locations) can be viewed as a Named Entity Recognition (NER) task, a well-established Natural Language Processing (NLP) task consisting of detecting and classifying remarkable subsets of a text or document~\cite{sang2003introduction}. In forensics fields, the kind of entity law enforcement agencies are interested in can change from case to case. However, when working in criminal cases, information like who committed or is being accused of a crime, or the agents and agencies involved in a case, are general and typically useful kind of information~\cite{lopez2026crimener}.

NER has a lot of work done in several fields. Initially, it was focused on news and general documents, but it grew rapidly to other areas such as biomedicine~\cite{song2021deep}. In relation to the crime domain, there are works that focus on the extraction of legal entities~\cite{au2022ner} and on Cyber Threat Intelligence~\cite{wang2022aptner}. However, these works revolve around legal texts or specific types of crime, and are not suitable for general crime extraction for day-to-day criminal cases.

One of the major drawbacks for NER in scenarios where annotated data is scarce is that the annotation process is very time consuming and expensive. To alleviate this issue, meta-learning paradigms have emerged and are the most common approach in these cases~\cite{li2020metaner}. Among all of them, zero- and few-shot learning has become the most relevant for NER tasks in almost all settings~\cite{li2020few,moscato2023few}. Zero-shot NER refers to a learning paradigm where models pretrained on a source domain with huge annotated data are expected to correctly label entity categories which have never been explicitly seen during training. On the other hand, Few-Shot NER refers to a learning paradigm where models are expected to correctly learn the label distribution of the whole target domain and classify correctly all entity types while being trained with very few examples for each target (usually between 1 and 10 examples per class). These meta-learning paradigms are even more important to consider in crime domains where agents need to address not only the scarcity of annotated data but also the generalization capabilities of models to unseen or barely seen crime entities. 

To address these issues, we present CrimeNER database (CrimeNER-db), a case study, and a database for NER on general crime-related domains with more than 1.5K manually annotated documents from U.S. Department of Justice's press notes and reports of real terrorist attacks from past years. We label each detected entity into a predefined set of entity types with two levels of granularity, namely coarse and fine-grained entity types, as done in previous works~\cite{ding2021few}. The coarse entity types defined cover basic information about crimes, e. g., the criminal who committed the offense. Fine-grained entity types are defined to better contextualize each of the coarse entity types extracted from the document, e.g. the typology of the criminal activity mentioned or the number of criminals that committed the crime. We assess the quality of CrimeNER-db with experiments using general NER models. We provide fully supervised baselines using several flat and nested NER common architectures to extract coarse- and fine-grained entities in one- and two-stage passes. We also provide zero- and few-shot experiments using generalist Large Language Models (LLMs) in order to provide a comprehensive evaluation for agents and researchers of their potential to extract crime-related information with no or few previous information.

The main contributions of this work are summarized as follows:

\begin{itemize}

    \item We present CrimeNER-db, a case-study and database for general crime-related NER. We manually annotated more than 1.5K documents extracted from the U.S. Department of Justice's press notes and from public reports of real terrorist attacks that happened during 2021, and further processed to generate the final annotated documents.

    \item We define a two-level hierarchy for entity types, coarse- and fine-grained, and classify each token as part of a coarse entity with its corresponding fine-grained entity type, or as a non-entity. We define $4$ coarse entity types with their corresponding fine-grained entity types, which sum up to a total $21$ fine-grained entity types.

    \item We present a detailed statistical analysis of CrimeNER-db for both coarse- and fine-grained entity types. This analysis covers general statistics of the database and specific statistics of coarse- and fine-grained entities separately.

    \item We validate the quality of our database with a fully supervised baseline using different configurations of general NER models to extract independently coarse- and fine-grained entity types, as well as one-stage coarse- and fine-grained entity extraction with common flat and nested NER models. We also performed zero- and few-shot experiments using generalist LLMs and compared the results with fully supervised baselines to assess the generalizability of these models in general crime-related scenarios.
    
\end{itemize}

A preliminary version of this article was published in \cite{lopez2026named}. This work significantly improves \cite{lopez2026named} in the following aspects:

\begin{itemize}
    \item We have extended the related works section with a review of the state-of-the-art in Entity Linking (EL) technologies, a task closely related with both NER and the research project in which this work is framed.
    \item We provide a larger and more detailed statistical analysis of the first version of CrimeNER-db.
    \item We have extended the baseline experiments of \cite{lopez2026named} with several architectures for flat NER and nested NER. We compared the results of the new architectures with the original two-stage coarse- and fine-grained entity extraction.
    \item The new experiments demonstrate the high quality and robustness of CrimeNER-db in flat and nested NER tasks for general criminal information extraction.
\end{itemize}

The remainder of the work is organized as follows. In Section~\ref{sec::related works} we review the state-of-the-art of NER in data-scarce domains, as well as NER and NLP approaches related to crime investigation and forensics. We show the annotation procedure, data processing and statistics of CrimeNER-db in Section~\ref{sec::database}. Section~\ref{sec::experiments} provides the experimental setup and results of our experiments. Finally, the conclusions and future research directions that arise from this work are discussed in Section~\ref{sec::conclusions}.

\section{Related Work}
\label{sec::related works}

\subsection{NER on Data-Scarce Domains}

Although early work and datasets for NER focused on domains with huge data like CoNLL'03~\cite{sang2003introduction} or OntoNotes~\cite{weischedel2022ontonotes} for the news and general domains respectively, recent work on NER has addressed the necessity of developing datasets and models on data-scarce domains~\cite{dao2025overcoming}. In this line, a great number of works have focused on Biomedical NER~\cite{song2021deep}, remarking well-established datasets like JNLPBA~\cite{collier2004introduction}. In other data-scarce domains, such as crime-related domains, most efforts have revolved around the creation of datasets for cybercrime NER~\cite{wang2024threat}. Due to the absence of these annotated data, previous work tackled this problem by using a few-shot learning setting in crime domains using LLMs~\cite{zhang2025seclmner}. Although some work has addressed the extraction of legal entities from legal documents such as court decisions~\cite{leitner2019fine} or case documents~\cite{naik2023legal}, there is still no dataset on general crime-related NER from real-world scenarios. Note that data scarcity in some data-scarce domains can be overcome to some extent by using synthetic data \cite{MELZI2024102322,Bio26macias} or domain adaptation from general information \cite{Fierrez-Aguilar2005_AdaptedMultimodal,gonzalo2026lora}, but there is still a strong need for curated data for crime research and investigation using NER such as the one contributed here.

\subsection{Entity Linking in Forensic Scenarios}

Entity Linking (EL) is traditionally framed as the task of disambiguating entity mentions in text and connecting them to their corresponding entries in a large open-domain knowledge base such as Wikipedia or Wikidata, with benchmarks like AIDA-CoNLL~\cite{hoffart2011robust} being central to the development of the field. However, applying EL to forensic and crime-related scenarios is particularly challenging, since entities of interest (e.g., suspects, victims, or criminal organizations) are sometimes not represented in knowledge bases or appear under aliases. Consequently, most existing work reformulates entity linking in forensic scenarios as an entity resolution or identity-linking problem, aiming to determine whether different mentions across documents correspond to the same real-world criminal actor rather than to a fixed knowledge base entry. This line of work has been particularly prominent in the human trafficking domain, where knowledge graphs constructed from online escort advertisements are used to link entities such as phone numbers, locations, and aliases across ads~\cite{szekely2015building, kejriwal2017knowledge}. Similarly, on darknet marketplaces, several works have tackled vendor identity linking by casting it as an authorship attribution or authorship aliasing problem that matches vendor accounts across different markets~\cite{saxena2023vendorlink, wang2024multi}, while other works extend this identity linking problem beyond the dark web itself, connecting dark web aliases to open-web identities~\cite{arabnezhad2020light}.

\subsection{Crime-related NER and NLP}

As a support technology, together with the use of biometrics \cite{Gonzalez-Rodriguez2005_Bayesian,2013PTomeFSI_FacialRegions,2017_Handbook_Forensic_Ram}, NLP is perhaps the most important tool for legal, security and crime analysis and investigation~\cite{2023_ICDAR_LLMs-TopicsPublicDocs_Pena,PENA-Layout,garg2023comprehensive,sarzaeim2023systematic}. Forensic NLP tasks require models trained in legal corpus related to the crimes to analyze. For that, some works have trained variants of BERT for tasks such as classification of political content~\cite{hu2022conflibert}. Recent works on Forensic NLP have also focused on event extraction in different settings such as mass-shooting events~\cite{ihugba2025knowledge} or dark web marketplaces for firearms~\cite{porlou2024optimizing}. However, there are few works addressing the analysis of general crimes as a NER task~\cite{lorentzen2026entity} due to the lack of enough well-annotated data on the topic.

\section{CrimeNER: Database}
\label{sec::database}

In this section, we describe the data acquisition and preprocessing that we followed when constructing CrimeNER-db, the annotation procedure and guidelines, and the statistics of CrimeNER-db.

\subsection{Data Acquisition and Preprocessing}
\label{subsec::data acquisition}

The main attraction of models and techniques designed and trained with crime data is the ability of law enforcement agencies to deploy them in real-world scenarios and take advantage of them in their daily duties. In order to achieve that, crime-related databases need to be annotated with documents that law enforcement agencies use or that at least share part of the technical vocabulary employed by them. In addition, the extraction of critical information from real-world news reports is of great use for these agencies, especially in recent times, where the need to analyze documents mentioning criminal activities is increasing. In order to achieve these objectives, we extracted the texts for CrimeNER-db from two sources: the United States Department of Justice's (DOJ) press notes and the Global Terrorism Database (GTD)~\cite{lafree2007introducing}.

The DOJ is the principal federal executive department tasked with the enforcement of federal law and the administration of justice within the United States. As part of its duties, the DOJ releases press releases on a daily basis, with topics varying widely between different types of crime offenses (drugs, cybercrime, or fraud, e.g.) to initiatives and projects of the Department itself. To extract the sentences for CrimeNER-db and annotate them, we used the Kaggle dataset ``Department of Justice 2009-2018 Press Releases''~\footnote{\url{https://www.kaggle.com/datasets/jbencina/department-of-justice-20092018-press-releases}}. This dataset contains the DOJ's press releases between 2009 and 2018, made by the different divisions in the Department regarding all the topics the DOJ is interested in. Due to the huge number of press releases (approximately 200K press releases) and in order to make the database affordable to annotate, we selected the press releases between 100 and 500 characters long. From this subset, after preprocessing and removing noisy text, we selected more than 1.2K documents. Although these documents had almost all of the entity types we were considering, we found that terrorist attacks and organizations, as well as locations outside the United States, were greatly misrepresented. To alleviate this issue, we added documents from GTD.

GTD is an open-source database released by the U. S. National Consortium for the Study of Terrorism and Responses to Terrorism (START) and the U. S. Department of Criminology and Criminal Justice in 2007 that gathers information about terrorist events around the world from 1970 and 2020, and subsequently an extension of the database was released with terrorist events that happened during 2021. The information of each event ranges from the region and country where it happened, the target of the attack and the perpetrator, or the number of causalities and other technical details. There are also reports, descriptions, and one or more news articles for each event. To maintain the structure of the documents similar to the ones we extracted from the DOJ's press releases and to keep the database as up to date as possible, we selected the events' descriptions of the 2021 GTD release. We applied the same preprocessing as with the DOJ's documents and selected reports between 100 and 500 characters long. From this subset, we selected around 300 documents to complete the CrimeNER-db construction, with the database comprising more than 1.5k documents.

\subsection{Annotation Guidelines and Procedure}
\label{subsec::annotation}

The primary goal of CrimeNER-db is to provide researchers with a fine-grained general crime dataset. In order to do that, based on previous work on fine-grained NER databases~\cite{ding2021few}, we define a two-level hierarchy for entity types, coarse and fine-grained, and classify each token as part of a coarse entity with its corresponding fine-grained entity type, or as a non-entity. The coarse entity types and their corresponding fine-grained types defined for CrimeNER-db are the following:

\begin{itemize}
    \item \textbf{Crime}: Refers to illegal activities performed by an individual or group of people that violate the laws and regulations of the country where they happened. Terrorist attacks and events are also considered as Crime entities. As fine-grained crime entity types, we consider: Terrorism, Fraud, Illegal traffic, Theft, Drug-related crimes, Homicides, Sexual crimes, Hate crimes, and Other crimes.
    \item \textbf{Actor}: refers to the individual, group of people, or organization that committed a crime or is accused of committing one. We distinguish between 4 fine-grained actor entity types: Criminal Person (which can be an individual or a non-organized group of people), Criminal Organization, Terrorist Person (which can be an individual or a non-organized group of people), and Terrorist Organization.
    \item \textbf{Agent}: Refers to individuals or organizations that act against criminal activities or who act to enforce the law against criminals. We also consider government officials and members of government bodies as this type of entity. We divide this entity type into three different fine-grained types: Law Enforcement, Government and Legal.
    \item \textbf{Logistic}: Refers to specific useful details for law enforcement agencies that are mentioned in the documents. The fine entity types that we consider in this case are: Location, GPE, Date, Weapons \& Explosives, and Money.
\end{itemize}

\begin{figure}[t!]
\centering
\includegraphics[scale=0.40]{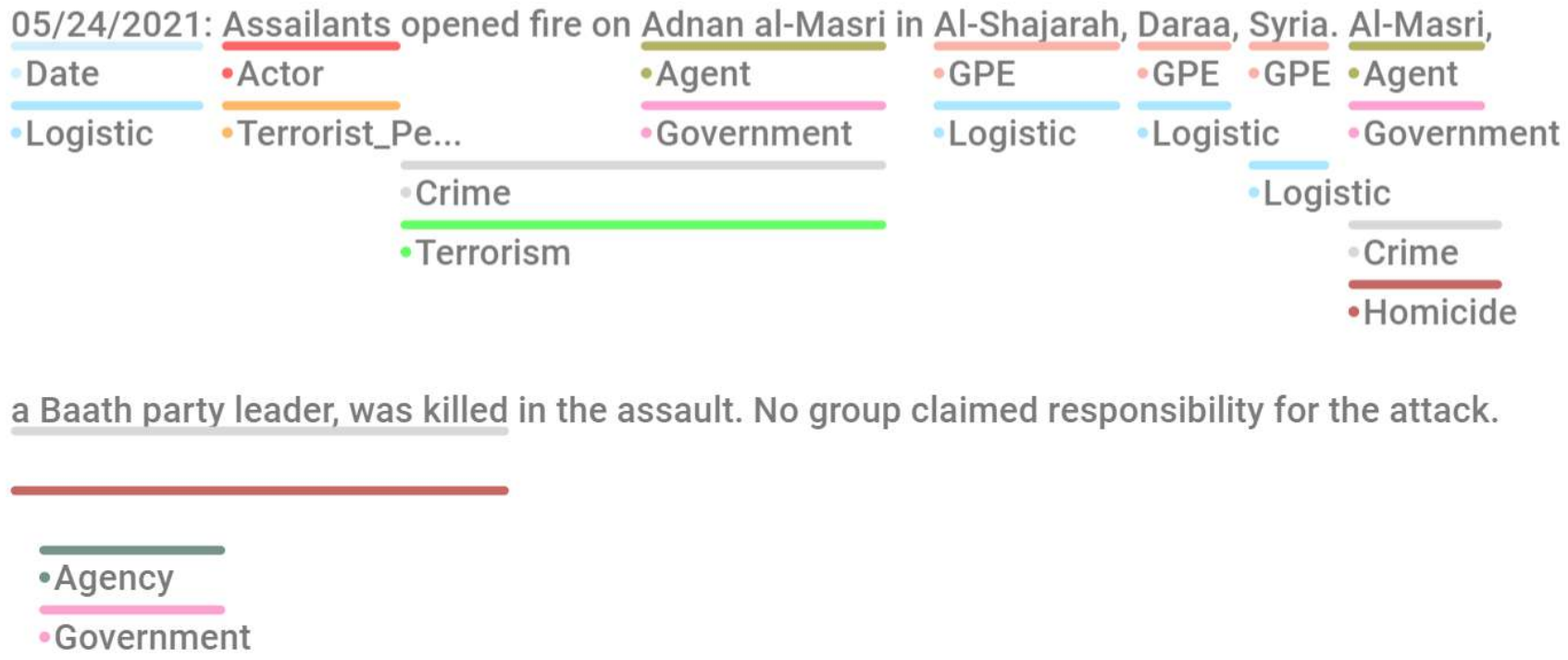}
\vspace{-10mm}
\caption{Example of an annotated document on Doccano with both coarse and fine-grained entity types and nested entities.} 
\label{fig::example_annotation}
\end{figure}

To better understand each coarse and fine entity label, we provide CrimeNER-db examples of all of them in Table~\ref{tab::examples}.

\subsubsection{Annotation Procedure}

The scientific description of the annotation and training procedures to derive intelligent systems applied to forensic and crime investigation has been crucial since the Daubert rule in 1993 \cite{Gonzalez-Rodriguez2005_Bayesian}. This scientific audit is now becoming a legal requirement \cite{irigoyen26risks} that attracts much attention and research and development efforts, especially around LLMs \cite{gonzalo2026cai,gonzalo2026lora}.

In the present work, to annotate the documents, we use the Doccano library~\cite{doccano}. We labeled every entity span as the corresponding coarse and fine entity type or left without marking if it was not an entity. There were some cases where some entities were nested inside others, and we annotated both of them, so CrimeNER-db is also suitable for Nested NER tasks. For example, in the text ``Attorney General for the District of Columbia'', we label the whole text as Agent-Legal and ``District of Columbia'', which is inside this entity, is also labeled as Logistic-GPE. We show an example of an annotated document with both coarse- and fine-grained entity types with nested entities in Fig.~\ref{fig::example_annotation}. During postprocessing we merged both entity types into the final Agent label.

The annotation process was carried out by $3$ independent annotators in three different rounds of annotation. In the first round, each annotator independently labeled the entities within each document. In the second round, each annotator reviewed and validated the entity annotations of the other annotators. If the discrepancies were larger than a certain margin threshold (especially with respect to the span length), the annotators would resolve the discrepancies jointly. In the final round, each annotator would independently perform a final sanity check on all entities, and any discrepancy would be discussed again jointly.

\subsection{CrimeNER-db Statistics}

\begin{table}[tph]
\centering
\setlength{\tabcolsep}{10pt} 
\caption{Statistics of CrimeNER database for each coarse entity type.}
\begin{tabular}{lrrr}
\hline
Entity type             & \# Entity spans & \# Tokens & \# Documents\\ \hline
Crime                   & 1341            & 6393      & 877         \\
Actor                   & 1421            & 2636      & 854         \\
Agent         & 1345            & 5910      & 683         \\
Logistic                & 2352            & 5039      & 1017         \\ \hline      
\end{tabular}
\label{tab::CrimeNER-db coarse stats}
\end{table}

\begin{figure}[tph]
  \centering
  \begin{tabular}{cc}
       \includegraphics[width=0.48\textwidth]{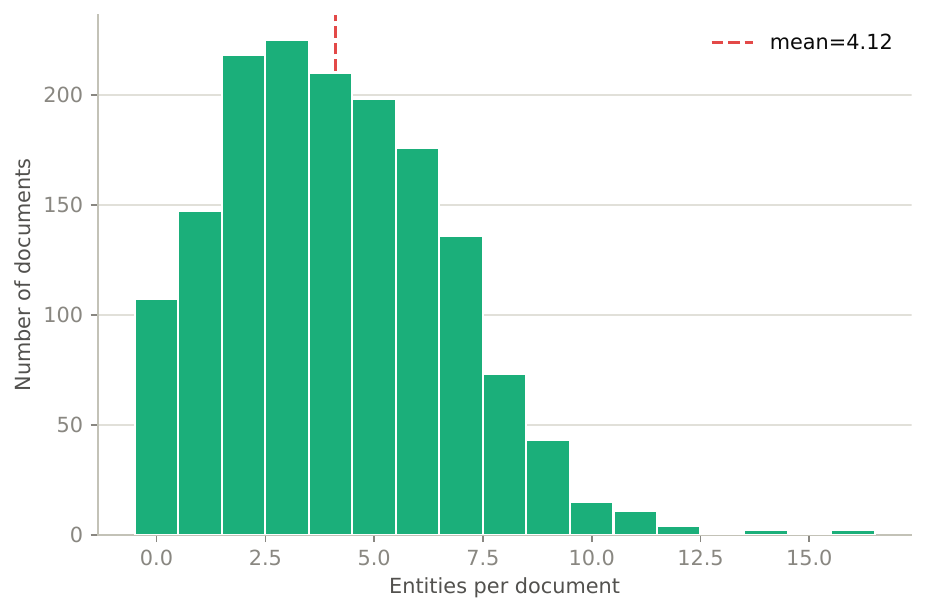} & 
       \includegraphics[width=0.48\textwidth]{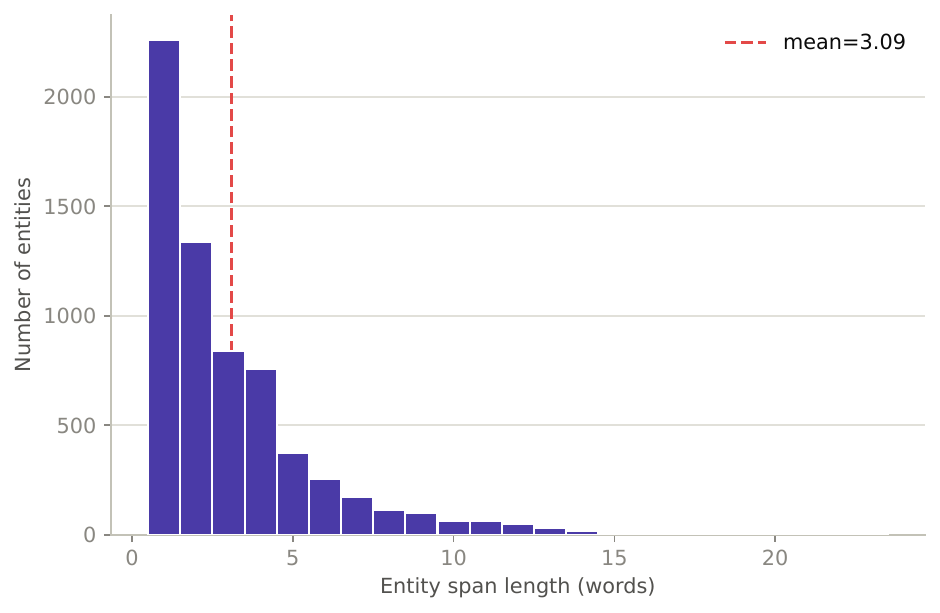}\\
       \footnotesize (a) & \footnotesize (b) 
  \end{tabular}
  \caption{Distribution of the number of entities per document (a) and the span length per entity (b) on CrimeNER-db. Mean number of entities per document and mean span length are respectively marked with a red dashed line.}
  \label{fig::entity distribution stats}
\end{figure}

\begin{figure}[tph]
  \centering
  \begin{tabular}{cc}
       \includegraphics[width=0.48\textwidth]{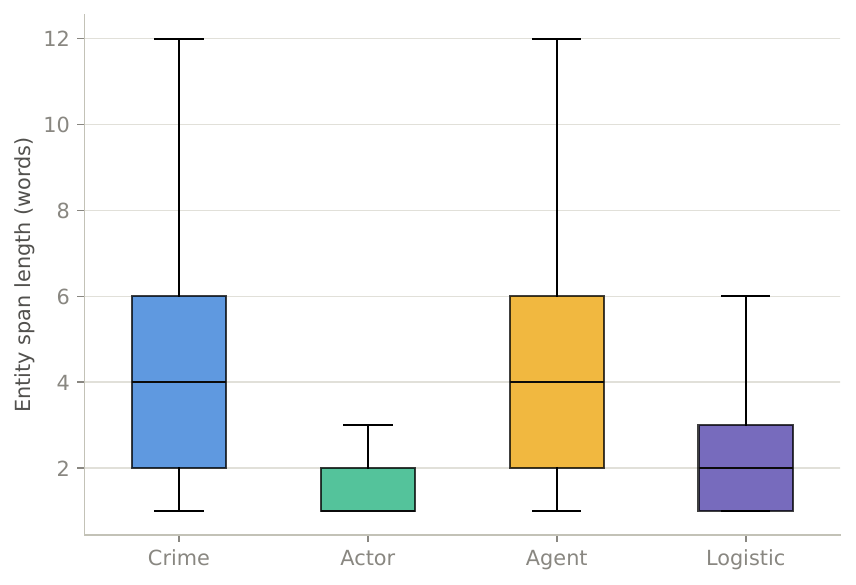} & 
       \includegraphics[width=0.48\textwidth]{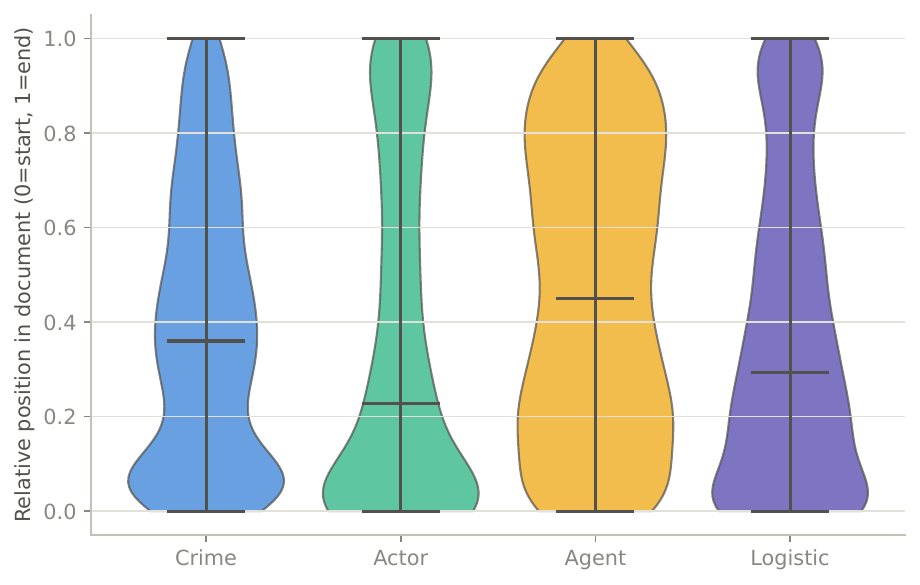}\\
       \footnotesize (a) & \footnotesize (b) 
  \end{tabular}
  \caption{Box plot of the span length (a) and violin plot of the span relative position (b) for each coarse entity type.}
  \label{fig::coarse span and violin stats}
\end{figure}

\begin{figure}[tph]
\centering
\includegraphics[width=0.95\textwidth]{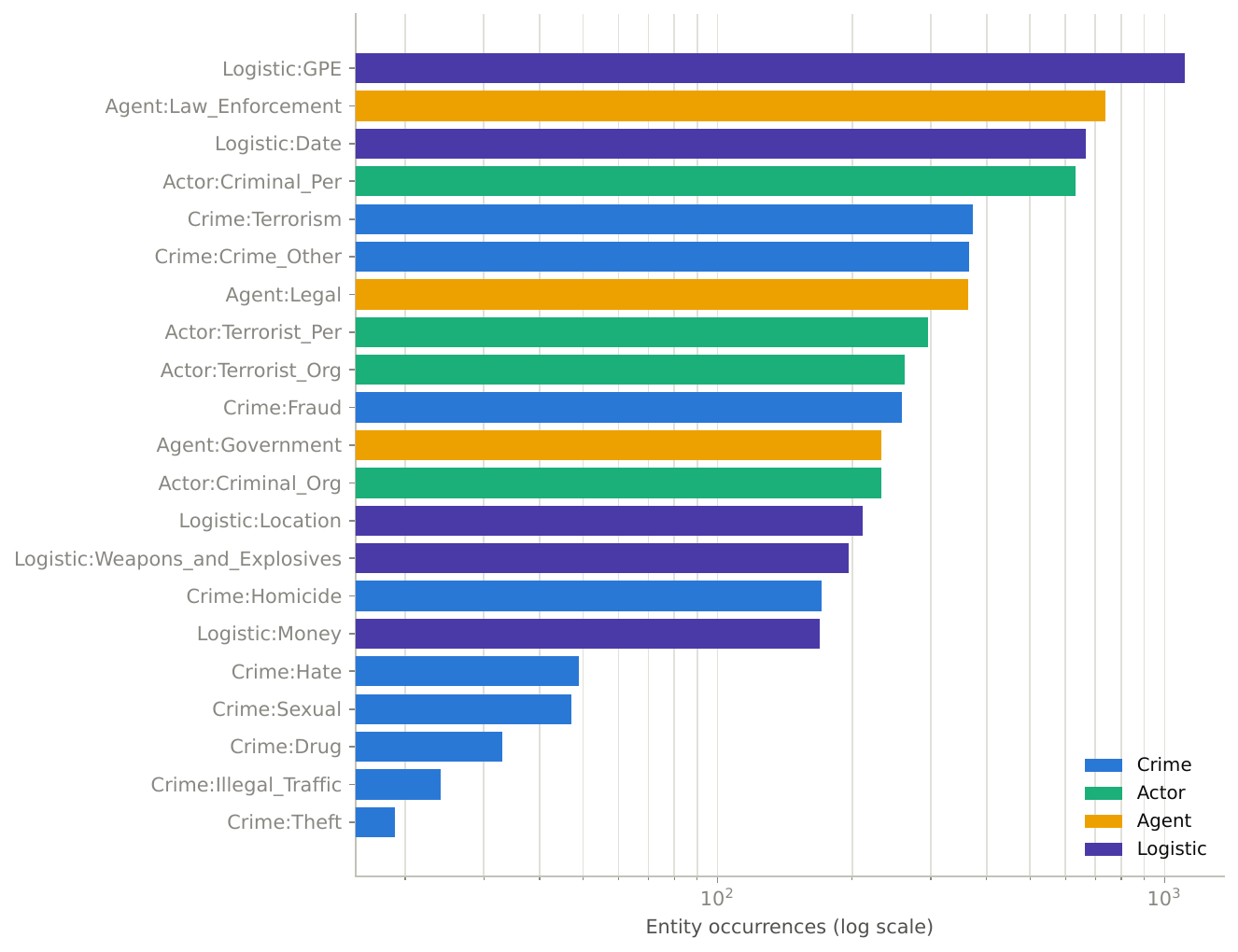}
\caption{Fine-grained entity span distribution of CrimeNER-db. PER and ORG refer to Person and Organization respectively.} 
\label{fig::span_distribution}
\end{figure}

After the whole annotation process and the cleaning of noisy or irregular documents, we ended up with a total of 1568 annotated documents from real-world sources. CrimeNER-db is, to the best of our knowledge, the first and largest data set for NER on general crime documents and terrorist reports. We report the CrimeNER-db statistics for each coarse entity type in Table~\ref{tab::CrimeNER-db coarse stats}. The most common entity label is Logistic, with $2352$ entity spans across all documents, while the least common one is the Crime, with only $1341$ entity spans, which is really close to the $1345$ entity spans of the Agent label. However, the entity type with the highest number of tokens is Crime. The number of documents in which each entity type appears is fairly balanced between the four classes.

The distribution of the number of entity spans per document is shown in Fig.~\ref{fig::entity distribution stats} (a). Most of the documents present between $2$ and $7$ named entities, with an average number of entities per document of $4.12$. Very few documents present more than $10$ entities, which is to be expected since the selected documents in CrimeNER-db are not lengthy enough in general to present that many entities. The distribution of the length of the entities is shown in Fig.~\ref{fig::entity distribution stats} (b). This distribution is heavily skewed to the left, with a large percentage of spans being formed by only one word. The mean span length in CrimeNER-db is $3.09$, and less than $100$ spans are formed by more than $10$ words.

The distribution of the span length for each coarse entity type is shown in Fig.~\ref{fig::coarse span and violin stats} (a). We observe that the most concentrated coarse label in terms of span length is the actor type, with a maximum span length of $3$ words. On the other hand, both both Crime and Agent labels show a larger deviation in span length, with both having a similar mean, maximum, and minimum span length. In Fig.~\ref{fig::coarse span and violin stats} (b) we show the violin plot of the relative position of the entity spans for each coarse entity label. Most of the coarse types are condensed at the start of the document and a little at the end, with few entities in the middle of the documents. The entity type that seems more balanced in terms of relative position is the Agent one, for which the violin plot, while thinner in middle positions of the document, is more balanced.

\section{Experiments}
\label{sec::experiments}

In this section, we report the experiments we performed on CrimeNER-db. To validate the quality of CrimeNER-db, we performed several experiments using common Pretrained Language Models (PLMs) in a fully supervised setting as an initial baseline. We performed experiments using common flat and nested NER architectures to provide a more robust evaluation of CrimeNER-db, as well as zero- and few-shot experiments using general open-source LLMs to assess their capabilities for general crime information extraction.

\subsection{Supervised Baseline}

\begin{table}[pt]
\centering
\setlength{\tabcolsep}{8pt}
\caption{Average Strict and Flexible F1 Scores on coarse and fine entity types after training on CrimeNER-db.\\}
\label{tab:f1_results}
\begin{tabular}{ccccc}
\toprule
 \textbf{Model} & \multicolumn{2}{c}{\textbf{Coarse}} & \multicolumn{2}{c}{\textbf{Fine}} \\
\cmidrule(lr){2-3} \cmidrule(lr){4-5}
 & \textbf{Strict} & \textbf{Flexible} & \textbf{Strict} & \textbf{Flexible} \\
\midrule
XLM-RoBERTa-Base~\cite{DBLP:journals/corr/abs-1911-02116}          & $0.714$ & $0.882$ & 
$0.597$ & $0.744$ \\
DeBERTa-V3-Base~\cite{he2021deberta} & $0.712$ & $\mathbf{0.889}$ & $0.605$ & $0.730$ \\
RoBERTa-Base~\cite{DBLP:journals/corr/abs-1907-11692}              & $\mathbf{0.725}$ & $0.888$ & $\mathbf{0.630}$ & $\mathbf{0.755}$ \\
AlBERT-Base-V2~\cite{DBLP:journals/corr/abs-1909-11942}            & $0.700$ & $0.867$ & $0.549$ & $0.714$ \\
DistilBERT-Base-Cased~\cite{Sanh2019DistilBERTAD}     & $0.686 $& $0.853$ & $0.546$ & $0.694$ \\
BERT-Base-Cased~\cite{DBLP:journals/corr/abs-1810-04805}           & $0.704$ & $0.873$ & $0.613$ & $0.729$ \\
\bottomrule
\end{tabular}
\end{table}

For supervised experiments, we selected $6$ common PLMs used for several NLP tasks. The selected models are XLM-RoBERTa-Base~\cite{DBLP:journals/corr/abs-1911-02116}, DeBERTa-V3-Base~\cite{he2021deberta}, RoBERTa-Base~\cite{DBLP:journals/corr/abs-1907-11692}, AlBERT-Base-V2~\cite{DBLP:journals/corr/abs-1909-11942}, DistilBERT-Base-Cased~\cite{Sanh2019DistilBERTAD} and BERT-Base-Cased~\cite{DBLP:journals/corr/abs-1810-04805}. For all of these experiments, we selected one of the models as the coarse entity extractor and another PLM (which may be the same as the coarse extractor) as the fine entity classifier. 

Given an input document, the coarse entity extractor extracts the coarse entity spans and classifies them into one of the coarse labels $4$. Then, the fine classifier takes the detected coarse spans with their labels and classifies them into the fine-grained types corresponding to each coarse label.

To evaluate the pipeline, we use two types of F1-score metric. The first is a strict F1 score where we only count a model prediction as correct if the detected entity span, the coarse label, and the fine label are the same as the ground truth annotation, with an offset threshold of $2$ that allows the span to take into account possible punctuation marks or white spaces  that the model predicts as parts of an entity and are not taken into account in the ground truth. The flexible F1 score also requires the predicted coarse and fine labels to be the same as the ground truth, but we consider a prediction correct if the predicted span overlaps or is a subset of the ground truth span. We considered this type of evaluation because the usual strict evaluation over penalizes small prediction errors and does not accurately reflect the model's performance.

The $6$ selected PLMs give us $36$ possible model configurations, since each model can be selected as a coarse entity extractor or fine entity classifier. For each model configuration, we trained both models jointly for $10$ epochs on a CrimeNER-db 80/10/10 train/val/test. We report the average evaluations of coarse entities of each model in each possible selection of the fine classifier and the average evaluations of fine entities of each model in each possible selection of coarse extractor in Table~\ref{tab:f1_results}. As expected, the flexible F1 score is more relaxed and provides better results. We observe that the best performing model on average with the strict coarse and the strict and flexible fine evaluation is RoBERTa-Base, while the best performing PLM on average with the flexible coarse evaluation is DeBERTa-V3-Base.

\subsection{Flat and Nested NER Experiments}

We compare the baseline against several flat and nested NER architectures. We selected both types of architecture because the number of tokens with more than $1$ coarse entity label (with the corresponding fine-grained label) corresponds to $7.7\%$ of entity tokens and the entity pairs that overlap across coarse types correspond to approximately the $8.2\%$ of all entity spans. This justifies the selection of flat and nested NER architectures, because the overlapping is present so nested entities need to be considered, but flat NER architectures are not structurally penalized too much for not detecting all nested entities.

\begin{table}[pt]
\centering
\setlength{\tabcolsep}{8pt}
\caption{Average Strict and Flexible F1 Scores on coarse and fine entity types for the selected flat NER architectures.\\}
\label{tab:flat_ner_results}
\begin{tabular}{ccccc}
\toprule
 \textbf{Model} & \multicolumn{2}{c}{\textbf{Coarse}} & \multicolumn{2}{c}{\textbf{Fine}} \\
\cmidrule(lr){2-3} \cmidrule(lr){4-5}
 & \textbf{Strict} & \textbf{Flexible} & \textbf{Strict} & \textbf{Flexible} \\
\midrule
BiLSTM-CRF~\cite{huang2015bidirectional}          & $0.652$ & $0.864$ & 
$0.501$ & $0.628$ \\
CNN-BiLSTM-CRF~\cite{ma2016end} & $0.653$ & $0.871$ & $0.508$ & $0.632$ \\
ID-CNN~\cite{strubell2017fast}             & $\mathbf{0.662}$ & $\mathbf{0.880}$ & $0.522$ & $0.660$ \\
Transformer + Softmax~\cite{devlin2019bert}            & $0.661$ & $0.876$ & $0.522$ & $\mathbf{0.735}$ \\
GLiNER~\cite{zaratiana2023gliner}     & $0.621 $& $0.804$ & $\mathbf{0.533}$ & $0.634$ \\
\bottomrule
\end{tabular}
\end{table}

\begin{table}[pt]
\centering
\setlength{\tabcolsep}{8pt}
\caption{Average Strict and Flexible F1 Scores on coarse and fine entity types for the selected nested NER architectures.\\}
\label{tab:nested_ner_results}
\begin{tabular}{ccccc}
\toprule
 \textbf{Model} & \multicolumn{2}{c}{\textbf{Coarse}} & \multicolumn{2}{c}{\textbf{Fine}} \\
\cmidrule(lr){2-3} \cmidrule(lr){4-5}
 & \textbf{Strict} & \textbf{Flexible} & \textbf{Strict} & \textbf{Flexible} \\
\midrule
Biaffine~\cite{yu2020named}          & $\mathbf{0.706}$ & $0.838$ & 
$0.530$ & $0.621$ \\
Stacked-CRF~\cite{ju2018neural} & $0.686$ & $\mathbf{0.870}$ & $0.524$ & $0.615$ \\
Constituency Tree~\cite{finkel2009nested}             & $0.705$ & $0.800$ & $0.544$ & $0.592$ \\
Seq2Seq~\cite{strakova2019neural}         & $0.535$ & $0.684$ & $0.232$ & $0.269$ \\
MRC~\cite{li2019unified}     & $0.702 $& $0.858$ & $\mathbf{0.611}$ & $\mathbf{0.715}$ \\
\bottomrule
\end{tabular}
\end{table}

The selected architectures for the flat NER experiments are: Iterated Dilated Convolutional Neural Networks (ID-CNN)~\cite{strubell2017fast}, the original token classification NER setup for BERT~\cite{devlin2019bert} which consists on Transformer blocks followed by Softmax functions, Bidirectional LSTM with Conditional Random Fields (BiLSTM-CRF)~\cite{huang2015bidirectional}, a combination of CNNs with the previous BiLSTM-CRF architecture (CNN-BiLSTM-CRF)~\cite{ma2016end}, and GLiNER~\cite{zaratiana2023gliner}, a generalist model for open NER tasks. These models were selected because they are the most common architectures for flat NER tasks in both closed and open domains without using LLMs.

All flat architectures except GLiNER share a common pretrained PLM backbone, in this case BERT-Base-Cased~\cite{DBLP:journals/corr/abs-1810-04805}. All models were trained for $5$ epochs with a learning rate of $3e-5$ with the same dataset split as the baseline. The results of the flat NER experiments are reported in Table~\ref{tab:flat_ner_results}.

The best performing architecture for coarse entity extraction is ID-CNN, with a strict and flexible F1 score of $0.662$ and $0.880$ respectively, followed for a small margin by the original implementation of BERT for token classification. For fine-grained entity extraction, the best performing architecture for the strict evaluation is GLiNER, and for the flexible evaluation, it is again BERT's original token classification setup. However, none of these architectures exceeds the supervised baselines in any evaluation.

The selected architectures and approaches for nested NER experiments are: Biaffine span classifier~\cite{yu2020named}, $3$ stacked CRF layers conditioned on the previous layer coverages (Stacked-CRF)~\cite{ju2018neural}, a Constituency Tree span classifier~\cite{finkel2009nested}, a Seq2Seq approach based on Facebook's BART\footnote{https://huggingface.co/facebook/bart-base} model~\cite{strakova2019neural} and a Machine Reading Comprehension (MRC) approach for NER~\cite{li2019unified}.

The nested NER architectures use BERT-Base-Cased~\cite{DBLP:journals/corr/abs-1810-04805} except for the Seq2Seq model that uses BART-Base. All models were trained for $5$ epochs with a learning rate of $3e-5$ with the same dataset split as the baseline. The results of the flat NER experiments are reported in Table~\ref{tab:nested_ner_results}.

In this case, the architecture with the best performance in the strict coarse evaluation is the Biaffine span classifier, with an F1 score of $0.706$, while in the flexible evaluation the best performing are the CRF layers stacked with an F1 score of $0.870$. The fine-grained entity extraction is dominated in both evaluations by the MRC framework. Again, none of these models surpass the supervised baselines, but they are closer than the flat NER architectures in terms of the strict evaluation. Nevertheless, in flexible evaluation for both coarse and fine-grained entities, flat NER architectures are closer to supervised baseline counterparts. 

\subsection{Zero- and Few-Shot Results}

\begin{figure}[pt]
\centering
\includegraphics[width=\textwidth]{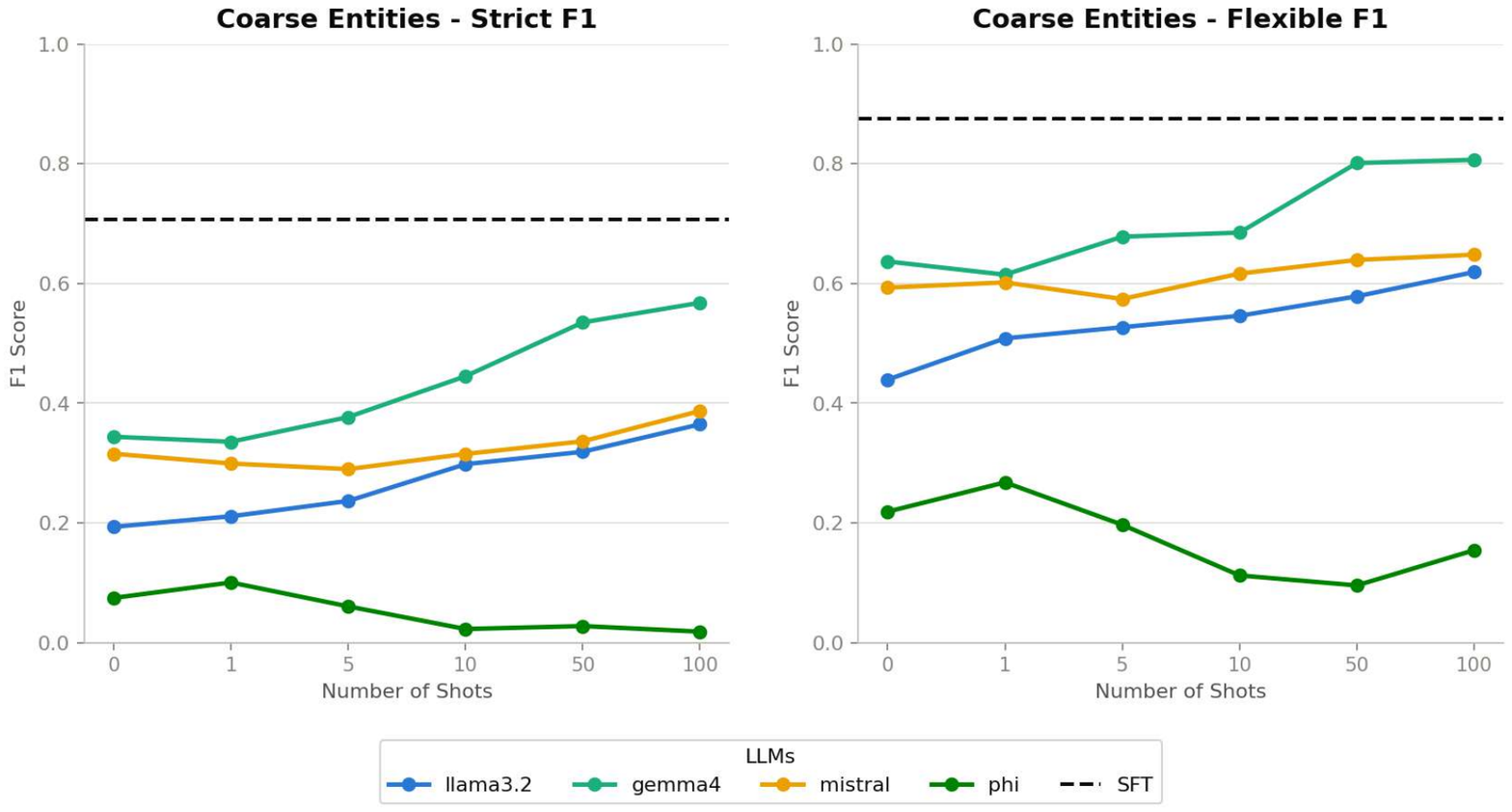}
\caption{Zero- and few-shot evaluation results using LLMs on coarse entities with strict (Left) and flexible (right) F1-scores. We compare the LLM results with the average performance of all $6$ baseline models as reported in Table~\ref{tab:f1_results}. \\
SFT = Supervised FineTuning (baseline)} 
\label{fig::zero-few-shot results}
\end{figure}

We acknowledge that criminal and forensic information extraction are fields where deployable models are required to perform well even with new unseen crimes. LLMs are known for their strong generalizability in domains different from those in which they were trained. To assess the capabilities of LLMs in the extraction of crime information, we provide an evaluation of $4$ open-source LLMs in zero- and few-shot settings. The selected LLMs are Llama3.2~\cite{grattafiori2024llama}, Gemma 4~\footnote{\url{https://deepmind.google/models/gemma/gemma-4/}}, Mistral 7B~\cite{Jiang2023Mistral7} and Phi-2\footnote{\url{https://ai.azure.com/catalog/models/microsoft-phi-2}}. We selected these models to assess the performance of small and large models from different families. We used Ollama\footnote{\url{https://ollama.com/}} for all LLM-based experiments.

For zero-shot experiments, we instruct each LLM to extract all coarse entity spans on the basis of some definitions we provide. For the few-shot experiments, we give the same instructions, but we provide $n$ examples of each coarse entity, where $n$ is the number of shots we provide in the instructions. For this round of experiments, the number of shots we consider are $1$, $5$, $10$, $50$, and $100$. In these experiments, we do not evaluate the fine entity extraction as some of the fine entities do not appear enough on CrimeNER-db to cover all the shots we are considering.

The results of our zero- and few-shot experiments are reported in Figure~\ref{fig::zero-few-shot results}. We compare the results with the best strict and flexible average F1-score reported for fully supervised training in Table~\ref{tab:f1_results}.

As expected, we observe a growing trend with both strict and flexible evaluations, where we achieve better results as more shots are considered, with the exception of Phi-2, which performs worse than any other LLM by far in every setting. We also observe that the difference in performance for all other LLMs with $50$ and $100$ shots is minimum, showing some kind of convergence between shots. We did not consider a larger number of shots as all LLMs showed a large performance decline when more than $100$ shots were considered, which may happen because the context of the model was overwhelmed~\cite{liu2024lost}.

As expected, LLMs do not outperform fully supervised baselines with few-shot experiments. This difference is more notable when comparing the strict F1 scores of the LLMs with the fully supervised baseline. This may be because LLMs are trained in an auto-regressive manner, which makes token and span classification tasks like NER difficult for them due to possible mismatches between the predicted and ground-truth spans. However, in a flexible evaluation, LLMs perform better, and while they do not outperform a fully supervised baseline, they achieve remarkable results.

In both types of evaluation, the best performing model is Gemma 4, which obtains the best flexible F1 score with $100$ shots. Although it does not outperform the fully supervised baseline for coarse entities, these results show that Gemma 4 is suitable for crime-related information extraction, even with very few samples of the crime being investigated.

These results show the quality and usefulness of CrimeNER-db as a dataset for NER tasks in forensic fields and crime-related analysis, in both fully supervised and few-shot settings.

\section{Conclusions and Future Work}
\label{sec::conclusions}

In this work, we present the CrimeNER database, a database for NER in forensic fields and general crime-related documents. CrimeNER-db is, to the best of our knowledge, the first database for NER in general crime-related real-world documents.

The documents for CrimeNER-db were extracted from real press notes of the U.S. Department of Justice and from real terrorist attack reports during 2021. After preprocessing and filtering these documents, we ended up with a total of 1568 annotated documents. These documents were annotated using a two-level hierarchy for entity types, coarse and fine-grained. The coarse entities defined in CrimeNER-db are: Crime, Actor, Agent \& Agency and Logistic. For each coarse entity type, we defined several fine-grained types, which in total sum up to 21 different entity types.

We validated the quality of CrimeNER-db with several experiments in a fully supervised setting using different PLMs as a supervised baseline. We compared this baseline with several common flat and nested NER architectures in a supervised setting and  against general LLMs in zero-shot and few-shot settings. 

Our experiments show that pretrained transformer-based models, finetuned in a two-stage fashion, perform better than any other typical supervised baseline for flat NER or nested NER. The PLMs that performed best on average were RoBERTa-Base and DeBERTa-V3-Base. The LLMs-based experiments, in zero- and few-shot settings, show that while generalist LLMs cannot outperform fully supervised PLMs in few-shot settings, they still perform well data-scarce criminal information extraction. However, the LLMs used in this study are rather small compared to current state-of-the-art models, which we will consider for future work, as they will probably perform better even in zero-shot settings.

For future work, several research directions remain open and will be explored. First, while CrimeNER-db is a large dataset for crime-related research, it is still small compared to other state-of-the-art datasets in NER. For this reason, we plan to extend CrimeNER-db with more documents from the same sources or even from sources in different languages, apart from English. For larger datasets, more sophisticated curation procedures, including semi-automatic annotation, may be required \cite{PENA-Layout}. We also plan to explore synthetic data generation to extend the database \cite{2026_arxiv_DocAI_DeAlcala}. The connection between criminal-related NER and existing forensic Entity Linking approaches is still underexplored; we will work on that research line in order to move towards a unified criminal Information Extraction framework using NLP approaches.

We will also explore multimodal architectures~\cite{Fierrez-Aguilar2005_BayesianAdaptation,2023_SNCS_Human-Centric_Pena,madikeri2025autocrime,miguel26icdar}, including combinations of NLP models such as those used here and visual models that process text images to improve the detection of criminal entities in multimodal settings such as documents \cite{lopez2025benchmarking,dealcala2026demo2}. Detecting AI-generated information \cite{2026_arxiv_DocAI_DeAlcala},
fakes \cite{MUNOZHARO2026103969}, manipulation \cite{pavel25iccv} and other risks associated with the use of intelligent systems \cite{irigoyen26risks} in NER-examined document repositories are also key topics on our agenda.

Finally, we believe that Named-Entity Recognition (NER) and Entity Linking (EL) technologies will be crucial for future crime investigation based on text and document repositories; but NER and EL applications go far beyond that. Among other applications, we are particularly interested in exploiting NER/EL for privacy and bias control in language intelligence systems \cite{pena2025addressing,mancera2025pba,mancera27mir}.

\begin{table*}[hbt!]
\centering
\small
\caption{Selected examples for every coarse and fine-grained entity type. PER and ORG refer to Person and Organization respectively.\\}
\label{tab::examples}
\rotatebox{0}{
\renewcommand{\arraystretch}{1} 
\setlength{\tabcolsep}{3.5pt}       

\begin{tabularx}{0.85\textheight}{|c|l|X|}
\hline
\textbf{Coarse} & \textbf{Fine-grained} & \textbf{Example} \\ \hline

\multirow{10}{*}{\textbf{Crime}}
 & Terrorism       & ... \blue{explosive device detonated targeting a military vehicle} in Qaya... \\ \cline{2-3}
 & Fraud           & ... one count of \blue{conspiracy to commit wire fraud}... \\ \cline{2-3}
 & Illegal Traffic & ... Credit Suisse \blue{illegally moved hundreds of millions of dollars} through... \\ \cline{2-3}
 & Theft           & ... and non-profit groups, \blue{stole more than a million dollars}... \\ \cline{2-3}
 & Drug-related crime            & ... one count of \blue{conspiracy to possess with intent to manufacture methamphetamine}... \\ \cline{2-3}
 & Homicide        & At least \blue{two employees were killed} in the attack... \\ \cline{2-3}
 & Sexual crime          & ... charged with multiple crimes involving \blue{sexual conduct with minors} in a foreign country... \\ \cline{2-3}
 & Hate crime            & ... commitment to ending \blue{unlawful lending discrimination}. \\ \cline{2-3}
 & Other crime           & \blue{Four women were kidnapped}, one person was killed... \\ \hline

\multirow{4}{*}{\textbf{Actor}}
 & Criminal (PER)  & In exchange for the bribes, \blue{Arroyo} and \blue{Becerril} devised a plan... \\ \cline{2-3}
 & Criminal (ORG)  & ... \blue{CCS} received net profits of approximately \$39.5 million through this scheme. \\ \cline{2-3}
 & Terrorist (PER) & \blue{Abdirahman J.}, an unaffiliated Jihadi-inspired extremist... \\ \cline{2-3}
 & Terrorist (ORG) & ... various terrorist organizations, including \blue{Ansar al-Islam}... \\ \hline

\multirow{3}{*}{\textbf{Agent}}
 & Law Enforcement & ... the United States \blue{Department of Justice} and the hard work of the federal prosecutors... \\ \cline{2-3}
 & Government      & ... August 2009 by the \blue{Dominican Office of the Prosecutor General}. \\ \cline{2-3}
 & Legal           & ... said \blue{Principal Deputy Assistant Attorney General Vanita Gupta} of the Civil Rights Division. \\ \hline

\multirow{5}{*}{\textbf{Logistic}}
 & Location               & ... protesters on the \blue{Cesar Gaviria viaduct} in Pereira, Risaralda, Colombia. \\ \cline{2-3}
 & GPE                    & ... detonated targeting a military vehicle in \blue{Qaya, Khanaqin district, Diyala, Iraq}. \\ \cline{2-3}
 & Date                   & ... in late summer or fall of \blue{2007}, Ortiz and other Latin King members... \\ \cline{2-3}
 & Weapons and Explosives & ... discovered and safely defused \blue{explosive devices} intended to target civilians... \\ \cline{2-3}
 & Money                  & ... supervised release and a fine of \blue{\$250,000}. \\ \hline

\end{tabularx}}
\end{table*}

\bibliographystyle{elsarticle-num} 
\bibliography{main}






\end{document}